\documentclass[11pt]{article}

\usepackage[utf8]{inputenc}
\usepackage[T1]{fontenc}
\usepackage{graphicx}
\usepackage{booktabs}
\usepackage{amsmath}
\usepackage{amssymb}
\usepackage[margin=2cm]{geometry}
\usepackage{caption}
\usepackage{subcaption}
\usepackage{hyperref}
\usepackage{authblk}
\usepackage{enumitem}
\usepackage{float}
\usepackage{tabularx}

\hypersetup{
    colorlinks=true,
    linkcolor=blue,
    citecolor=blue,
    urlcolor=blue
}

\title{\textbf{MariSat: A Maritime Dataset for Instance Segmentation of Objects in Satellite and Aerial Images}}

\author[1]{Amir Abbes}
\author[1]{Ines Harrabi}
\author[1]{Lucas Justin Yirepoa Kinda}
\author[2]{Rim Trabelsi}
\author[3]{Adnane Cabani}
\author[1]{Fatma Abdelkefi}

\affil[1]{University of Carthage, SUP'COM, LR11TIC05, MEDIATRON, 2083, Ariana, Tunisia}
\affil[2]{University of Gabes, Hatem Bettaher IResCoMath Laboratory, Gabes, Tunisia}
\affil[3]{University of Rouen Normandie, ESIGELEC, IRSEEM, 76000 Rouen, France}

\date{}

\begin{document}

\maketitle

\begin{abstract}
\noindent
Automated maritime surveillance from satellite and aerial imagery requires large, precisely annotated datasets, which remain scarce for the instance-segmentation task, particularly for small vessels in cluttered port environments. We present MariSat, a new benchmark dataset of 1260 aerial and satellite images covering diverse port and coastal scenes, annotated at the pixel level for eight maritime object classes (sailboat, yacht, jet-ski, fishing boat, cruise ship, military vessel, tugboat and cargo ship). The dataset was produced through a semi-automatic annotation pipeline combining the text-promptable segmentation model SAM~3 with a cascade of geometric and colorimetric post-processing filters, followed by a manual correction and quality-control pass performed with the CVAT annotation platform. We describe the image-collection methodology, the annotation and correction process, and the resulting data organization. We also report class-wise statistics for the training, validation, and test splits. MariSat has already been used to fine-tune and benchmark segmentation and detection models (SAM~3 and YOLO11) for real-time maritime monitoring. We report detailed quantitative and per-class results for both tasks. The MariSat dataset is publicly available on GitHub : https://github.com/amirabbes/P2M-Maritime-Segmentation

\vspace{0.3cm}
\end{abstract}
\vspace{0.5cm}

% ============================================================
\section{Introduction}
% ============================================================

More than 90\% of world trade is carried by sea, making the automated monitoring of coasts and ports a strategic problem for maritime safety and traffic management. Existing operational solutions, such as the Automatic Identification System (AIS) and coastal radar, suffer from well-known limitations: AIS only reports vessels that are large enough or willing to transmit their position, leaving small or non-cooperative craft invisible, while radar struggles to classify vessel types and degrades in adverse weather. Traditional computer vision approaches, in turn, lack robustness to the diversity of vessel shapes and textures, as well as to visual clutter caused by wave foam and water glare.

To address these limitations, recent research has increasingly turned to deep learning architectures incorporating attention mechanisms. For instance, the work in \cite{link1}, based on multi-head self-attention, demonstrated that this approach can significantly enhance object detection performance in the maritime domain by better capturing contextual relationships between vessels and their surroundings, while their subsequent work systematically evaluated attention-based models on maritime datasets, showing consistent improvements in mean average precision \cite{link2}. These approaches are particularly relevant for overcoming the visual clutter caused by wave foam and water glare, as attention modules help suppress complex background noise and enhance key instance features.

Beyond architectural innovations, the scarcity of annotated maritime imagery has spurred the development of specialized data augmentation techniques. The authors in \cite{link3} proposed an object-centric contour-aware augmentation method using superpixels of varying granularity, which preserves object shape integrity while generating diverse training samples from limited data.

These recent advances offer a way to both detect and precisely delineate maritime objects from overhead imagery, enabling fine-grained vessel classification in addition to detection. Training and evaluating such models, however, requires image-level and pixel-level annotated data that is costly to produce manually, especially for small objects that occupy only a handful of pixels in high-altitude imagery. This scarcity of richly annotated, class-diverse maritime instance-segmentation data motivated the creation of the dataset described in this paper.

MariSat targets two related tasks: (i) multi-class instance segmentation of maritime objects in individual aerial/satellite images, and (ii) object detection (bounding-box localization and classification) of the same eight vessel categories, which we derive from the segmentation masks. The remainder of this paper details the data-construction methodology (Section~\ref{sec:curation}), reports detailed segmentation and detection experiments (Section~\ref{sec:experiments}), and discusses ethical considerations and limitations (Sections~\ref{sec:ethics} and~\ref{sec:limitations}).

% ============================================================
\section{Dataset Construction}
\label{sec:curation}

%===========================================================

\subsection{Image collection}

\begin{figure}[H]
    \centering
    \includegraphics[width=\linewidth]{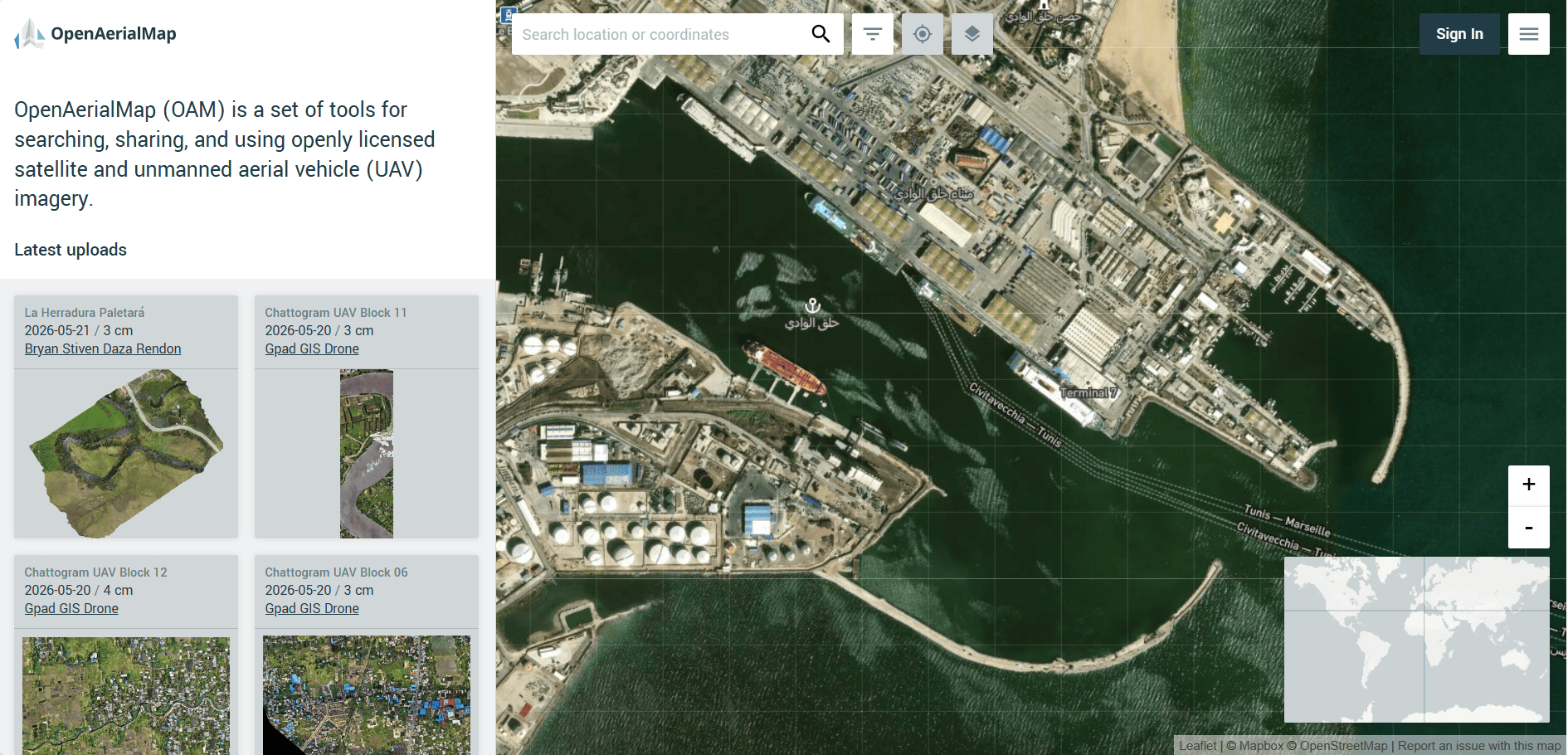}
    \caption{OpenAerialMap platform interface}
    \label{fig:oam}
\end{figure}

Images were collected from OpenAerialMap (OAM) \cite{oam}, an open repository of freely licensed aerial and satellite imagery. As shown in Figure~\ref{fig:oam}, the OpenAerialMap platform provides access to high-resolution orthophotos, which were selected for this study because they enable the reliable identification of small maritime objects across geographically diverse port and coastal environments.

\begin{figure}[H]
    \centering
    \includegraphics[width=\linewidth]{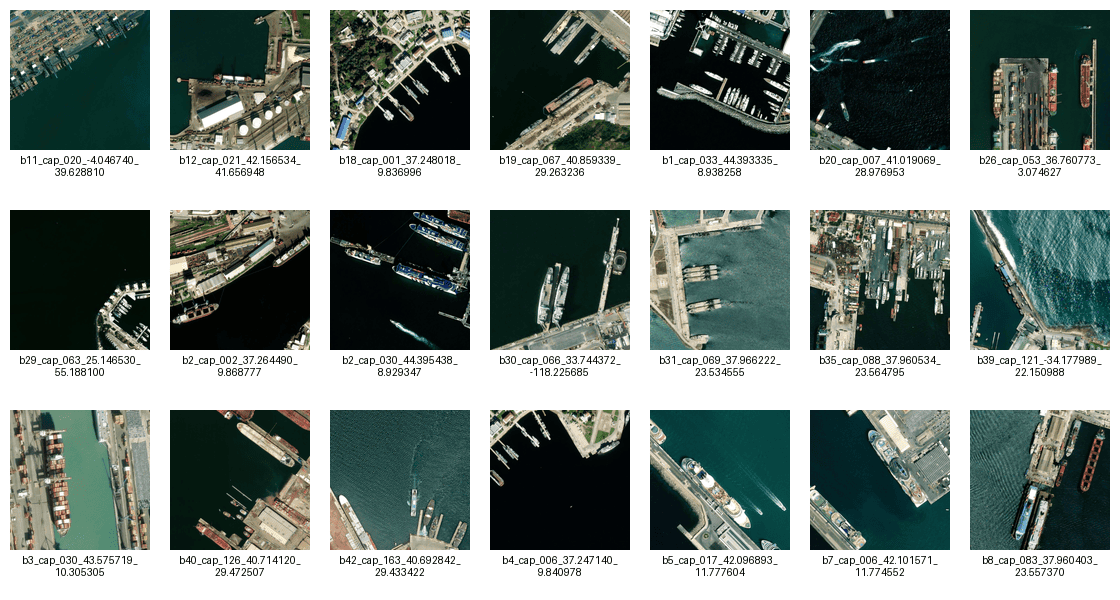}
    \caption{Representative sample of images collected for MariSat}
    \label{fig:samples}
\end{figure}

Data acquisition was automated using a Python script based on the OpenAerialMap API. Following an iterative process, the operator specifies a geographic starting point (GPS coordinates) and a movement direction; the script then traverses image tiles along the selected axis and downloads the corresponding areas. This pipeline produced a final collection of 1260 images, encompassing a wide variety of maritime scenes and covering all target classes. A representative sample is shown in Figure~\ref{fig:samples}.

\subsection{Semi-automatic annotation with SAM~3}

Manually annotating 1260 images at pixel level was prohibitively time-consuming. We therefore built a semi-automatic annotation pipeline around SAM3SemanticPredictor, the text-promptable variant of the SAM~3 segmentation model \cite{sam3}. Each raw image is passed to the model together with eight free-text prompts, one per target class, and the model returns instance segmentation masks together with their predicted class. The eight classes and their associated prompts are listed in Table~\ref{tab:prompts}.

\begin{table}[H]
\centering
\caption{Maritime object classes and the text prompts used for semi-automatic annotation with SAM~3}
\label{tab:prompts}
\renewcommand{\arraystretch}{1.2}

\begin{tabularx}{1\textwidth}{p{3.2cm}X}
\toprule
\textbf{Class} & \textbf{Text prompt} \\
\midrule
Sailboat & aerial view sailboat with mast on water \\
Yacht & aerial view luxury motor yacht on water \\
Jet-ski & aerial view jet ski personal watercraft on water \\
Fishing boat & aerial view fishing boat trawler commercial harbor on water \\
Cruise ship & aerial view large white cruise ship passenger vessel port \\
Military vessel & aerial view grey military warship frigate navy vessel \\
Tugboat & aerial view tugboat short wide red orange hull port \\
Cargo ship & aerial view cargo ship bulk carrier flat deck freight port \\
\bottomrule
\end{tabularx}
\end{table}

\begin{figure}[H]
    \centering

    \begin{subfigure}{0.32\linewidth}
        \centering
        \includegraphics[width=\linewidth,height=4cm,keepaspectratio]{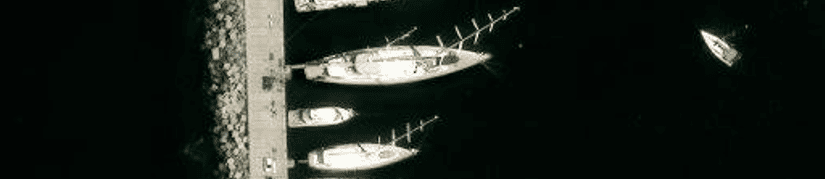}
        \caption{Sailboat}
    \end{subfigure}
    \begin{subfigure}{0.32\linewidth}
        \centering
        \includegraphics[width=\linewidth,height=4cm,keepaspectratio]{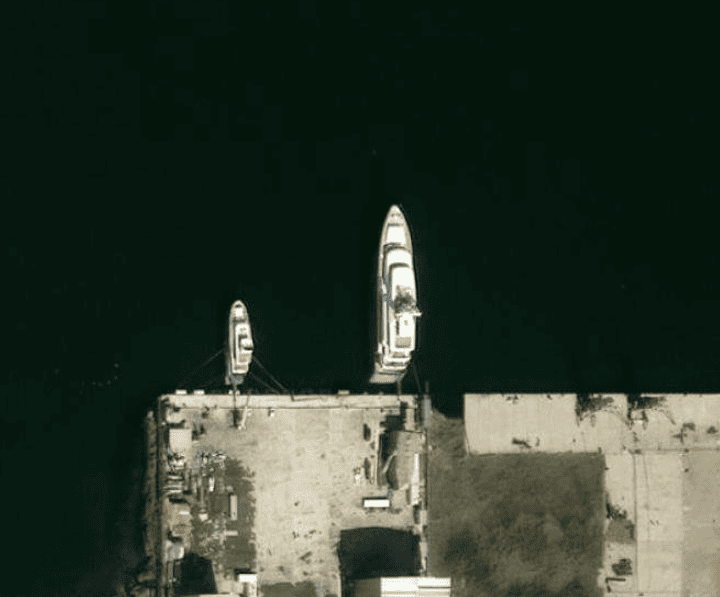}
        \caption{Yacht}
    \end{subfigure}
    \begin{subfigure}{0.32\linewidth}
        \centering
        \includegraphics[width=\linewidth,height=4cm,keepaspectratio]{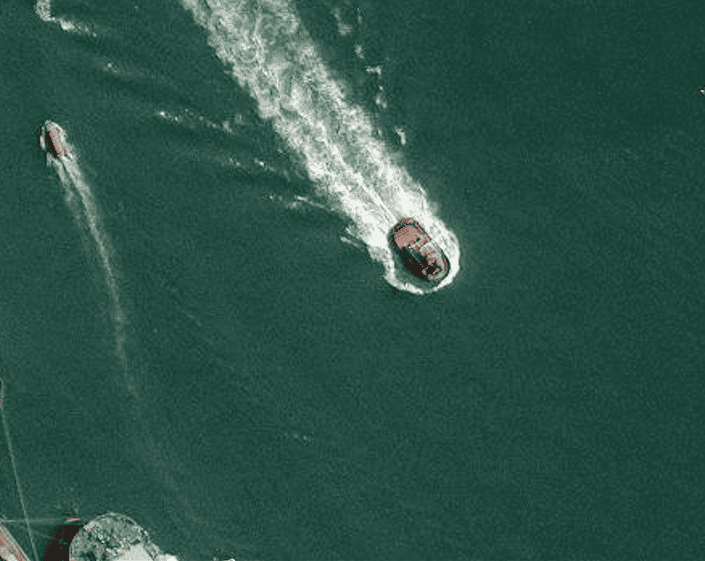}
        \caption{Tugboat}
    \end{subfigure}

    \vspace{0.2cm}

    \begin{subfigure}{0.32\linewidth}
        \centering
        \includegraphics[width=\linewidth,height=4cm,keepaspectratio]{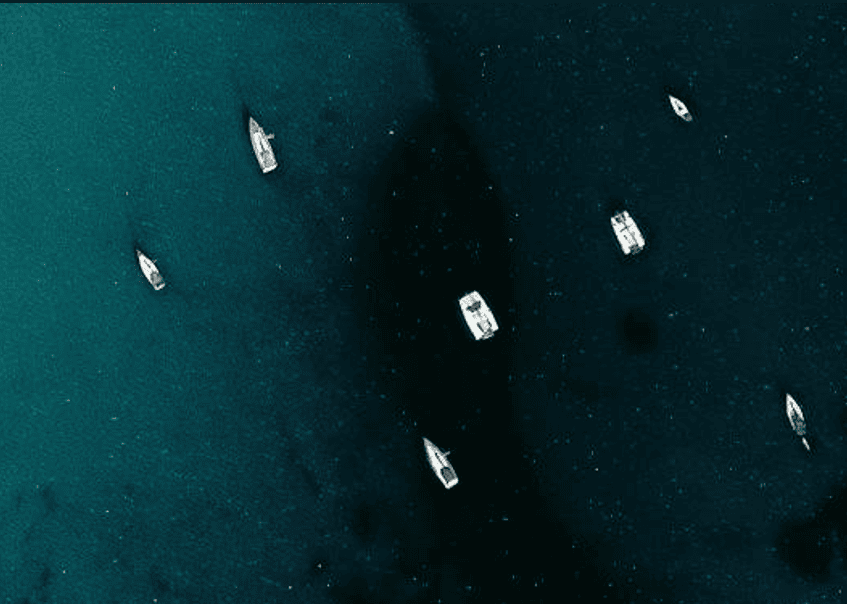}
        \caption{Fishing boat}
    \end{subfigure}
    \begin{subfigure}{0.32\linewidth}
        \centering
        \includegraphics[width=\linewidth,height=4cm,keepaspectratio]{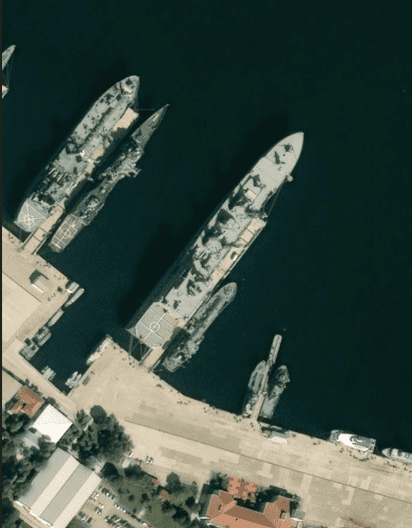}
        \caption{Military vessel}
    \end{subfigure}
    \begin{subfigure}{0.32\linewidth}
        \centering
        \includegraphics[width=\linewidth,height=4cm,keepaspectratio]{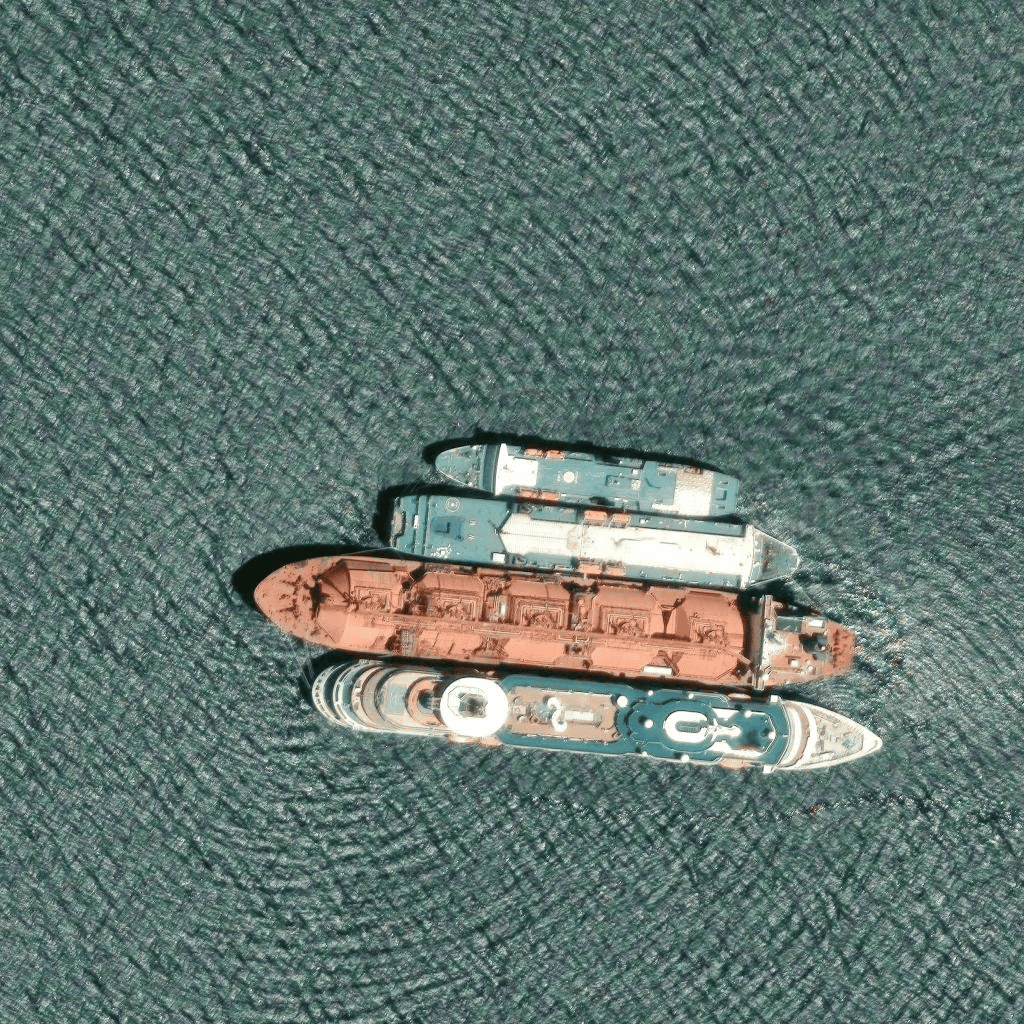}
        \caption{Cargo ship}
    \end{subfigure}

    \caption{Representative examples of six maritime object classes}
    \label{fig:classes}
\end{figure}

The Figure~\ref{fig:classes} illustrates representative samples from different maritime object classes included in the dataset.

\subsection{Post-processing filters}

Raw SAM~3 predictions contain false positives and spurious masks. Four filters were applied in cascade to clean the pre-annotations before manual review:

\begin{itemize}[leftmargin=*]
    \item \textbf{Non-maximum suppression (NMS):} when two masks overlap with an IoU greater than 50\%, only the most confident one is kept
    \item \textbf{Size filter:} masks covering more than 30\% of the image surface are discarded, as empirical observations showed that large masks almost always correspond to mis-segmented sea background
    \item \textbf{Shape filter:} masks with an oriented length-to-width ratio below 1.2 are discarded, since vessels typically exhibit elongated shapes irrespective of their orientation in the image
    \item \textbf{Water filter:} detections located outside the maritime surface are rejected through colorimetric analysis in the Hue-Saturation-Value (HSV) color space, which separates chromatic information (hue and saturation) from brightness, making water-region identification less sensitive to illumination variations
\end{itemize}

\begin{figure}[H]
    \centering

    \begin{subfigure}{0.95\linewidth}
        \centering
        \includegraphics[width=\linewidth,height=0.25\textheight,keepaspectratio]{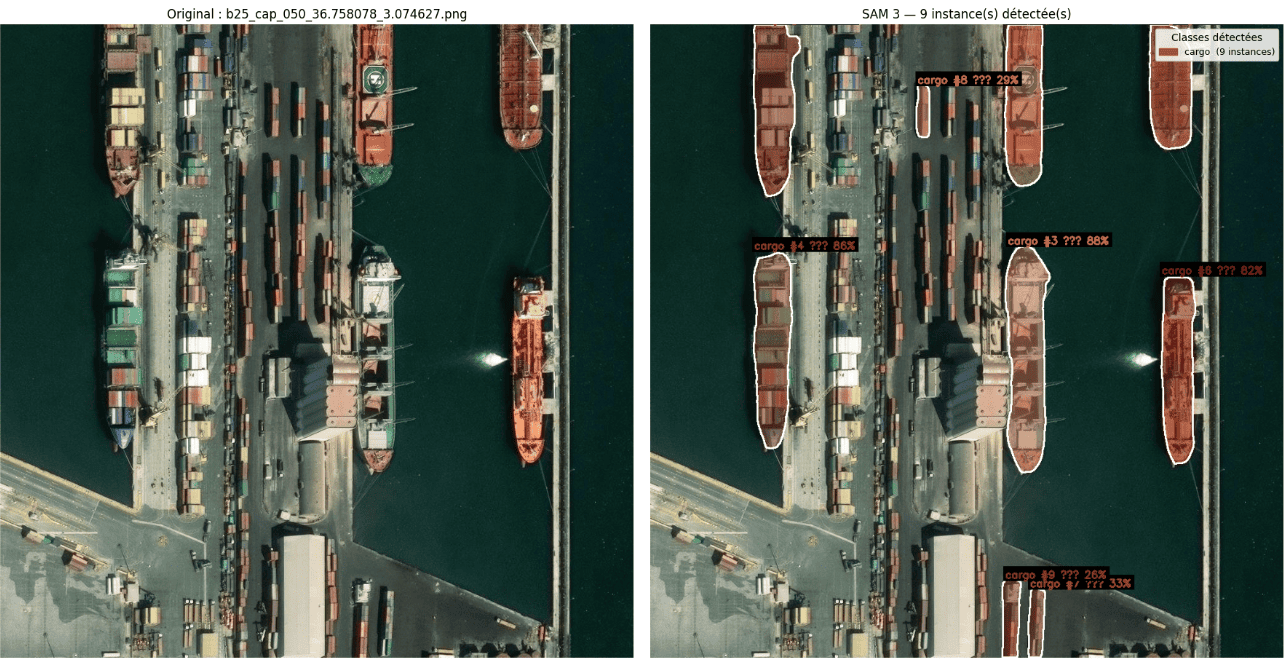}
        \caption{Correct segmentation and classification}
    \end{subfigure}

    \vspace{0.15cm}

    \begin{subfigure}{0.95\linewidth}
        \centering
        \includegraphics[width=\linewidth,height=0.25\textheight,keepaspectratio]{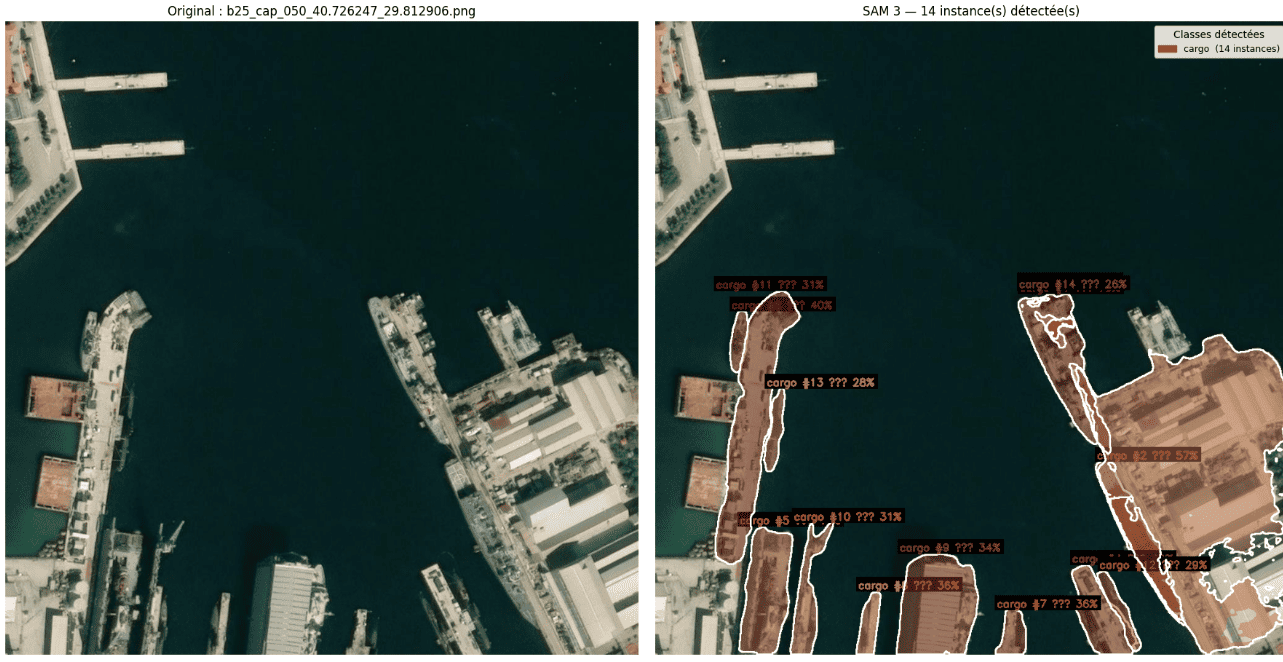}
        \caption{Presence of false positives}
    \end{subfigure}

    \vspace{0.15cm}

    \begin{subfigure}{0.95\linewidth}
        \centering
        \includegraphics[width=\linewidth,height=0.25\textheight,keepaspectratio]{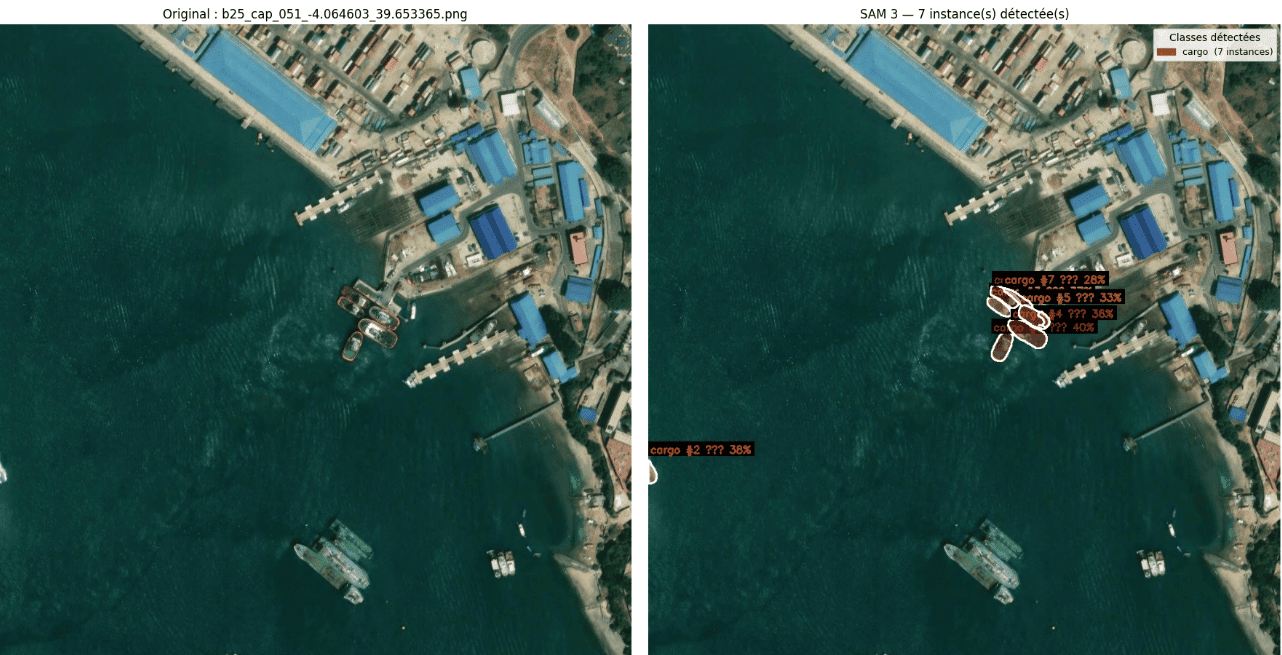}
        \caption{Presence of false negatives}
    \end{subfigure}

    \caption{Representative results of the SAM~3 pre-annotation process}
    \label{fig:preanno}
\end{figure}

The Figure~\ref{fig:preanno} illustrates three representative cases obtained during the pre-annotation stage with SAM~3, highlighting both the strengths of the model and the limitations that motivated the subsequent manual correction phase performed using CVAT \cite{cvat}.

\subsection{Manual correction with CVAT}

\begin{figure}[H]
    \centering
    \includegraphics[width=\linewidth]{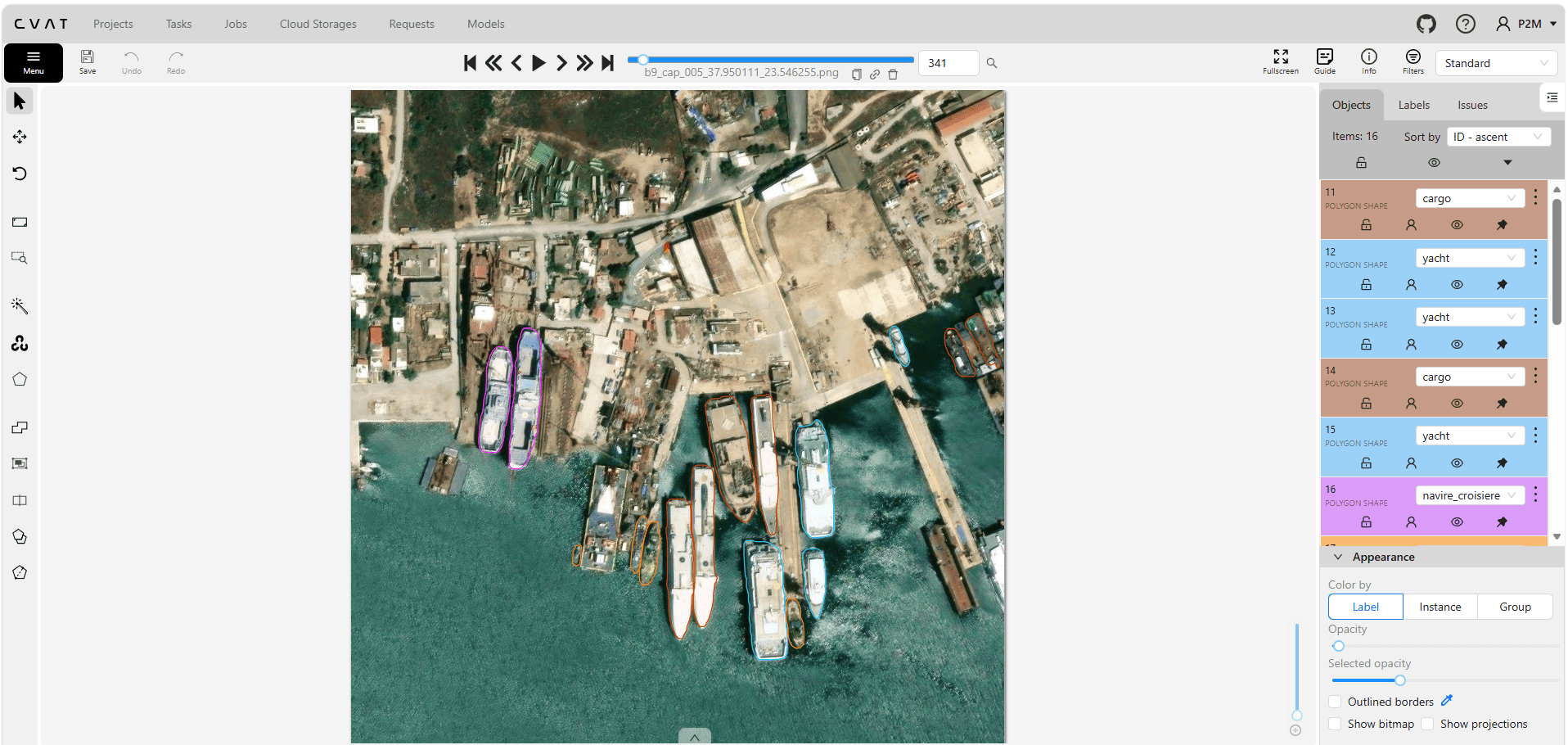}
    \caption{CVAT interface used for the manual correction}
    \label{fig:cvat}
\end{figure}

As shown in Figure~\ref{fig:cvat}, the filtered annotations are exported in XML format and imported into the CVAT annotation platform \cite{cvat} for manual refinement and quality control. Three types of corrections were systematically applied by the annotators:

\begin{itemize}[leftmargin=*]
    \item removal of residual false positives not caught by the automatic filters;
    \item addition of objects missed by SAM~3, typically small or partially occluded vessels;
    \item refinement of imprecise or poorly delineated mask contours.
\end{itemize}

\subsection{Dataset organization}
\label{sec:organization}

The corrected annotations are organized in batches of 20 images. Each batch folder contains the raw PNG images together with a CVAT XML file encoding the segmentation polygons of every annotated object. Each polygon is associated with a class label and the pixel coordinates of its contour.

Table~\ref{tab:dist_before} reports the initial per-class instance distribution across the training, validation and test splits, before any rebalancing was applied. Several classes, most notably fishing boat, yacht and cargo ship, are substantially over-represented relative to sailboat, jet-ski and cruise ship, and the initial per-split proportions deviate from the targeted 70\%/15\%/15\% train/validation/test ratio for several classes.

\begin{table}[H]
\centering
\caption{Instance distribution across splits before rebalancing}
\label{tab:dist_before}
\small
\begin{tabular}{lrrrr}
\toprule
\textbf{Class} & \textbf{Train} & \textbf{Val} & \textbf{Test} & \textbf{Total} \\
\midrule
sailboat & 626 & 6 & 179 & 811 \\
yacht & 4069 & 252 & 1049 & 5370 \\
jet-ski & 909 & 23 & 328 & 1260 \\
fishing boat & 5918 & 708 & 1424 & 8050 \\
cruise ship & 370 & 90 & 108 & 568 \\
military vessel & 778 & 59 & 190 & 1027 \\
tugboat & 1493 & 516 & 467 & 2476 \\
cargo ship & 2726 & 431 & 592 & 3749 \\
\bottomrule
\end{tabular}
\end{table}

To bring the per-class split proportions closer to this target and to reduce downstream training bias, a batch-level rebalancing step was applied, reassigning entire annotation batches between splits so as to better equalize per-class train/validation/test ratios without breaking image--batch integrity. Table~\ref{tab:dist_after} reports the resulting distribution, used for all experiments reported in Section~\ref{sec:experiments}.

\begin{table}[H]
\centering
\caption{Instance distribution across splits after batch-level rebalancing}
\label{tab:dist_after}
\small
\begin{tabular}{lrrrr}
\toprule
\textbf{Class} & \textbf{Train} & \textbf{Val} & \textbf{Test} & \textbf{Total} \\
\midrule
sailboat & 626 & 154 & 140 & 920 \\
yacht & 4074 & 579 & 923 & 5576 \\
jet-ski & 909 & 181 & 170 & 1260 \\
fishing boat & 5922 & 1454 & 1088 & 8464 \\
cruise ship & 385 & 76 & 107 & 568 \\
military vessel & 794 & 125 & 190 & 1109 \\
tugboat & 1595 & 446 & 440 & 2481 \\
cargo ship & 2790 & 397 & 566 & 3753 \\
\bottomrule
\end{tabular}
\end{table}

Rebalancing brings every class markedly closer to the 70/15/15 target, most notably for sailboat (from 1\% to 17\% validation share) and tugboat (from 21\% to 18\% validation share), while leaving the total number of images per split unchanged.

The annotated dataset is split into training, validation and test image sets, following the strategy summarized in Table~\ref{tab:split}.

\begin{table}[H]
\centering
\caption{Dataset split into training, validation and test subsets}
\label{tab:split}
\begin{tabular}{lrr}
\toprule
\textbf{Split} & \textbf{Batches} & \textbf{Images} \\
\midrule
Train & 42 & 840 \\
Validation & 10 & 200 \\
Test & 11 & 220 \\
\midrule
\textbf{Total} & \textbf{63} & \textbf{1260} \\
\bottomrule
\end{tabular}
\end{table}

MariSat comprises 1260 images collected across geographically diverse port and coastal areas, annotated with pixel-accurate instance masks for eight maritime object classes. As detailed in Section~\ref{sec:organization} (Tables~\ref{tab:dist_before} and~\ref{tab:dist_after}), class frequencies are imbalanced, with cargo ships, yachts, and fishing boats dominating the dataset, while sailboats, jet-skis and cruise ships are comparatively under-represented. This imbalance was mitigated downstream through batch re-balancing across splits (Section~\ref{sec:organization}) and the use of class-weighted training losses when the dataset was employed for model fine-tuning.

Two complementary usages of the dataset were validated during its construction: instance segmentation, using the pixel-level polygon annotations directly, and object detection, using bounding boxes derived from the same polygon annotations. The dataset was used to fine-tune and benchmark SAM~3 \cite{sam3} and YOLO11 \cite{yolo11} for these two tasks respectively, detailed experimental setup and results are reported in Section~\ref{sec:experiments}.

% ============================================================
\section{Experiments and Results}
\label{sec:experiments}
This section reports the fine-tuning setup and results for both models trained on MariSat: SAM~3 for instance segmentation and YOLO11 (nano and large variants) for object detection. We present the experimental setup and training hyperparameters, followed by overall and per-class results on the test set. 
% ============================================================

\subsection{Experimental setup}

\textbf{Instance segmentation:} SAM~3 \cite{sam3} was fine-tuned using its Perception Encoder text encoder \cite{pe} and SAM~2 image encoder \cite{sam2}, together with a lightweight text-guided detector head trained on the MariSat training set. Class re-balancing (Section~\ref{sec:organization}) and regularization were applied to mitigate the residual class imbalance. For evaluation purposes, the three most visually similar and smallest-footprint classes, namely sailboat, jet-ski and fishing boat, were grouped into a single ``small vessel'' category, yielding a 6-class evaluation setting (small vessel, yacht, cruise ship, military vessel, tugboat, cargo ship).

\textbf{Object detection:} YOLO11 \cite{yolo11} was fine-tuned in its nano (YOLO11n) and large (YOLO11l) variants on bounding boxes derived from the segmentation polygons. Table~\ref{tab:hparams} reports the training hyperparameters used for each variant, and Table~\ref{tab:augment} reports the data augmentation techniques applied during training.

\begin{table}[H]
\centering
\caption{Training hyperparameters}
\label{tab:hparams}
\small
\begin{tabular}{lrr}
\toprule
\textbf{Parameter} & \textbf{YOLO11n} & \textbf{YOLO11l} \\
\midrule
Epochs & 150 & 150 \\
Batch size & 16 & 6 \\
Patience & 50 & 50 \\
Dropout & -- & 0.2 \\
Learning rate (lr0) & 0.01 & 0.01 \\
Weight decay & 0.0005 & 0.0005 \\
\bottomrule
\end{tabular}
\end{table}

\begin{table}[H]
\centering
\caption{Data augmentation techniques applied}
\label{tab:augment}
\small
\begin{tabular}{lr}
\toprule
\textbf{Augmentation} & \textbf{Value} \\
\midrule
Horizontal flip (fliplr) & 30\% \\
Rotation (degrees) & $\pm 5^{\circ}$ \\
Mixup & 5\% \\

\bottomrule
\end{tabular}
\end{table}

\subsection{Quantitative results}

Table~\ref{tab:overall} reports the overall performance of both models on the held-out test set.

\begin{table}[H]
\centering
\caption{Overall performance on the MariSat test set}
\label{tab:overall}
\begin{tabular}{llccccc}
\toprule
\textbf{Model} & \textbf{Task} & \textbf{Precision} & \textbf{Recall} & \textbf{mIoU} & \textbf{mAP50} & \textbf{mAP50:95} \\
\midrule
SAM~3 & Instance Segmentation & 0.888 & 0.818 & 0.809  & -- & -- \\
YOLO11n & Detection & 0.612 & 0.465 & -- & 0.461 & 0.326 \\
YOLO11l & Detection & 0.519 & 0.589 & -- & 0.528 & 0.389 \\
\bottomrule
\end{tabular}
\end{table}

SAM~3 achieves the strongest overall precision and segmentation quality, consistent with its text-promptable design being closely aligned with the semi-automatic annotation pipeline. Between the two YOLO11 variants, the large model trades precision for a markedly higher recall (+0.124) and a better mAP50:95 (+0.063) than the nano model, reflecting the classic accuracy/efficiency trade-off relevant for deployment in real-time maritime monitoring: the nano model is more conservative (fewer false positives, more missed detections), while the large model detects more true vessels at the cost of more false positives.

\subsection{Per-class results}

\textbf{SAM~3:} Table~\ref{tab:sam3_class} reports per-class precision, recall, F1-score, pixel accuracy and mIoU on the 6-class evaluation setting described above.

\begin{table}[H]
\centering
\caption{Per-class results for SAM~3 on the MariSat test set}
\label{tab:sam3_class}
\begin{tabular}{lccccc}
\toprule
\textbf{Class} & \textbf{Precision} & \textbf{Recall} & \textbf{F1} & \textbf{Accuracy} & \textbf{IoU} \\
\midrule
small vessel (sailboat + jet-ski + fishing boat) & 0.927 & 0.689 & 0.789 & 0.829 & 0.638 \\
yacht & 0.923 & 0.947 & 0.935 & 0.967 & 0.747 \\
cruise ship & 0.858 & 0.773 & 0.813 & 0.978 & 0.972 \\
military vessel & 0.894 & 0.721 & 0.797 & 0.962 & 0.801 \\
tugboat & 0.837 & 0.812 & 0.824 & 0.916 & 0.803 \\
cargo ship & 0.887 & 0.966 & 0.925 & 0.891 & 0.893 \\
\midrule
\textbf{Macro average} & \textbf{0.888} & \textbf{0.818} & \textbf{0.847} & \textbf{0.924} & \textbf{0.809} \\
\bottomrule
\end{tabular}
\end{table}

\begin{figure}[H]
    \centering
    \includegraphics[width=0.97\linewidth]{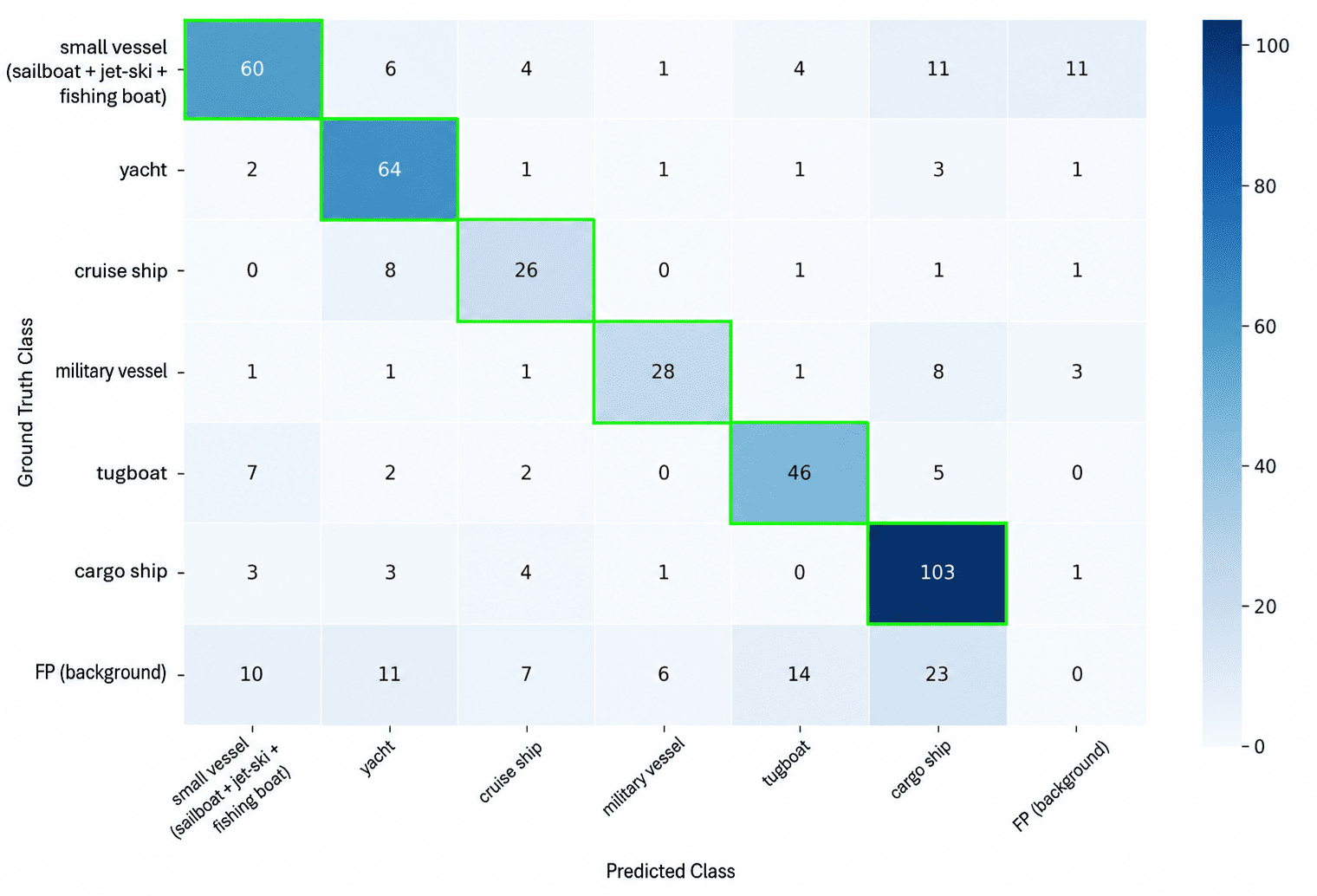}
    \caption{Confusion matrix for the fine-tuned SAM~3 model on the MariSat test set (6-class setting)}
    \label{fig:confusion}
\end{figure}

The small-vessel category shows the lowest recall (0.689) of all classes, consistent with the class-imbalance and small-object-size challenges discussed in Section~\ref{sec:organization} : sailboats, jet-skis and small fishing boats occupy few pixels in aerial imagery and are more easily missed by the segmentation head. Figure~\ref{fig:confusion} reports the corresponding confusion matrix, which shows that most misclassifications occur between the small-vessel category and cargo ship / tugboat, and that a non-trivial number of predictions are pure false positives (23 for cargo ship, 14 for tugboat), reflecting residual background confusion despite the post-processing filters of Section~\ref{sec:curation}.

\textbf{YOLO11:} Table~\ref{tab:yolo_n} and Table~\ref{tab:yolo_l} report per-class detection results for the nano and large variants, respectively, using the original 8-class taxonomy.

\begin{table}[H]
\centering
\caption{Per-class results for YOLO11n on the MariSat test set}
\label{tab:yolo_n}
\begin{tabular}{lrrccc c}
\toprule
\textbf{Class} & \textbf{Images} & \textbf{Instances} & \textbf{Precision} & \textbf{Recall} & \textbf{mAP50} & \textbf{mAP50:95} \\
\midrule
sailboat & 21 & 179 & 0.522 & 0.235 & 0.269 & 0.140 \\
yacht & 67 & 990 & 0.564 & 0.588 & 0.605 & 0.404 \\
jet-ski & 11 & 111 & 1.000 & 0.000 & 0.126 & 0.063 \\
fishing boat & 67 & 1325 & 0.376 & 0.164 & 0.201 & 0.115 \\
cruise ship & 31 & 62 & 0.461 & 0.790 & 0.562 & 0.471 \\
military vessel & 33 & 143 & 0.613 & 0.713 & 0.636 & 0.505 \\
tugboat & 55 & 252 & 0.709 & 0.567 & 0.640 & 0.431 \\
cargo ship & 111 & 550 & 0.651 & 0.662 & 0.647 & 0.481 \\
\midrule
\textbf{All} & \textbf{220} & \textbf{3612} & \textbf{0.612} & \textbf{0.465} & \textbf{0.461} & \textbf{0.326} \\
\bottomrule
\end{tabular}
\end{table}

\begin{table}[H]
\centering
\caption{Per-class results for YOLO11l on the MariSat test set}
\label{tab:yolo_l}
\begin{tabular}{lrrccc c}
\toprule
\textbf{Class} & \textbf{Images} & \textbf{Instances} & \textbf{Precision} & \textbf{Recall} & \textbf{mAP50} & \textbf{mAP50:95} \\
\midrule
sailboat & 21 & 179 & 0.439 & 0.585 & 0.453 & 0.258 \\
yacht & 67 & 990 & 0.580 & 0.701 & 0.662 & 0.489 \\
jet-ski & 11 & 111 & 0.383 & 0.081 & 0.180 & 0.099 \\
fishing boat & 67 & 1325 & 0.364 & 0.441 & 0.315 & 0.199 \\
cruise ship & 31 & 62 & 0.389 & 0.726 & 0.491 & 0.413 \\
military vessel & 33 & 143 & 0.634 & 0.797 & 0.680 & 0.542 \\
tugboat & 55 & 252 & 0.687 & 0.687 & 0.743 & 0.567 \\
cargo ship & 111 & 550 & 0.678 & 0.696 & 0.704 & 0.541 \\
\midrule
\textbf{All} & \textbf{220} & \textbf{3612} & \textbf{0.519} & \textbf{0.589} & \textbf{0.528} & \textbf{0.389} \\
\bottomrule
\end{tabular}
\end{table}

Across both variants, jet-ski is consistently the hardest class to detect (mAP50:95 of 0.063 for YOLO11n and 0.099 for YOLO11l), which is consistent with it being both the smallest object class and one of the least represented in the training set (Table~\ref{tab:dist_after}). YOLO11n reaches a perfect precision but zero recall on jet-ski, indicating that the model essentially never fires on this class within its confidence threshold rather than confusing it with another class. Moving from the nano to the large variant substantially improves recall for small and under-represented classes, with sailboat recall increasing from 0.235 to 0.585 and fishing boat recall from 0.164 to 0.441, at a moderate cost in precision for these same classes. This confirms that model capacity partially compensates for the class imbalance, but does not fully resolve it for the rarest class, jet-ski.

% ============================================================
\section{Conclusion}
% ============================================================

We introduced MariSat, a dataset of 1260 aerial and satellite images annotated with pixel-level instance masks for eight maritime object classes, built through a semi-automatic pipeline combining SAM~3 \cite{sam3} text-promptable pre-annotation, a cascade of geometric and colorimetric filters, and a manual correction pass in CVAT \cite{cvat}. The dataset covers diverse port and coastal scenes and is organized into training, validation, and test splits suitable for both instance-segmentation and object-detection benchmarks. Baseline experiments with SAM~3 \cite{sam3} and YOLO11 \cite{yolo11} demonstrate the usefulness of our proposed dataset.

% ============================================================
\section{Ethics Statement}
\label{sec:ethics}
% ============================================================

All source imagery originates from OpenAerialMap \cite{oam}, an open, freely licensed catalogue of satellite imagery; no personally identifiable information is captured by the dataset, which depicts vessels and port infrastructure rather than individuals. The dataset is intended for research purposes related to maritime object detection and segmentation.

% ============================================================
\section{Limitations}
\label{sec:limitations}
% ============================================================

The dataset comprises only 1260 images, with class imbalance, particularly for jet-skis and sailboats. This imbalance negatively affects detection performance for under-represented classes. Although annotations were refined using CVAT \cite{cvat}, minor inaccuracies may remain. Finally, the use of imagery from a single source, OpenAerialMap \cite{oam}, may limit generalization to other sensors and resolutions.

\subsection*{Data Availability}

To foster reproducible research and future developments in maritime vision, the complete dataset and annotations are publicly available at https://github.com/amirabbes/P2M-Maritime-Segmentation.

% ============================================================
% REFERENCES
% ============================================================


\begin{thebibliography}{9}
\bibitem{link1}
W. Messaoud, R. Trabelsi, A. Cabani, et F. Abdelkefi, « Multi-Head Self Attention for Enhanced Object Detection in the Maritime Domain », in The IEEE 22th International Conference on Cyberworlds (CW2023), 2023.

 
\bibitem{link2}
W. Messaoud, R. Trabelsi, A. Cabani, et F. Abdelkefi, « Maritime object detection using attention mechanism », SIViP 18, 1833–1845 (2024) \url{https://doi.org/10.1007/s11760-023-02897-1}

 
\bibitem{link3}
F. Dornaika, D. Sun, K. Hammoudi, J. Charafeddine, A. Cabani, et C. Zhang, « Object-centric contour-aware data augmentation using superpixels of varying granularity », Pattern Recognition, vol. 139, p. 109481, 2023.

\bibitem{oam}
OpenAerialMap, ``A set of tools for searching, sharing and using openly licensed satellite and UAV imagery.'' \url{https://openaerialmap.org}

\bibitem{pe}
D.~Bolya, P.~Misra, et al., ``Perception Encoder: The best visual embeddings are not at the output of the network,'' arXiv:2504.13181 (2025).

\bibitem{sam2}
N.~Ravi, V.~Gabeur, Y.-T.~Hu, et al., ``SAM~2: Segment Anything in Images and Videos,'' arXiv:2408.00714 (2024).


\bibitem{cvat}
CVAT.ai Corporation, \textit{Computer Vision Annotation Tool (CVAT)} (2023). \url{https://github.com/cvat-ai/cvat}



\bibitem{sam3}
Ultralytics, ``SAM 3: Segment Anything with Concepts,'' Ultralytics Documentation, 2025. \url{https://docs.ultralytics.com/models/sam-3}.



\bibitem{yolo11}
G.~Jocher and J.~Qiu, \textit{Ultralytics YOLO11}, version 11.0.0, Ultralytics (2024). \url{https://github.com/ultralytics/ultralytics}

\end{thebibliography}
\end{document}